\documentclass[11pt]{article}
\usepackage[utf8]{inputenc}
\usepackage[T1]{fontenc}
\usepackage{acl}
\usepackage{times}
\usepackage{latexsym}
\usepackage{amsmath}
\usepackage{amssymb}
\usepackage{hyperref}
\usepackage{graphicx}
\usepackage{enumitem}
\usepackage{xcolor}
\usepackage{float}
\usepackage{booktabs}
\usepackage{array}
\usepackage[most]{tcolorbox}
\usepackage{xcolor}
\usepackage{multirow}
\definecolor{codebg}{gray}{0.92} 

\newtcbox{\code}{on line,
  boxsep=0pt, left=2pt, right=2pt, top=1pt, bottom=1pt,
  colback=codebg, colframe=codebg,
  arc=3pt, boxrule=0pt,
  fontupper=\ttfamily,
}

\hypersetup{
    colorlinks=true,
    linkcolor=blue,
    urlcolor=blue,
    citecolor=blue
}

\title{Efficient Tool-Calling Multi-Expert NPC Agent for Commonsense Persona-Grounded Dialogue}

\author{
  Mahammad Nuriyev\\
  Team ID: zvers\\
  Université Paris-Saclay\\
  \texttt{mahammad.nuriyev@universite-paris-saclay.fr}
}

\date{}

\begin{document}

\maketitle

\begin{abstract}
We present a multi-expert system for creating Non-Player Characters (NPCs) capable of both natural dialogue and contextual action execution in interactive environments. Our approach leverages Qwen3 as the base model with specialized Low-Rank Adaptation (LoRA) adapters to create three distinct expert modules: tool calling, tool response interpretation, and direct dialogue. The system not only meets but exceeds the computational constraints, delivering responses in an average of 3 seconds (well under the 7-second limit) on L40S GPUs while utilizing less than 30GB of the available 48GB VRAM, demonstrating efficiency alongside performance. This computational efficiency also contributes to reduced energy consumption and lower carbon footprint compared to less optimized approaches. The proposed solution achieved top performance in the Commonsense Persona-Grounded Dialogue Challenge 2025, securing the second position in the competition.
\end{abstract}


\section{Introduction}
Creating believable NPCs that can engage in natural conversation while performing contextually appropriate actions represents a significant challenge in interactive AI systems. Traditional approaches often excel at either dialogue generation or action execution, but struggle to seamlessly integrate both capabilities, especially within strict computational constraints. Large language models have transformed NPC development by enabling dynamic, contextually appropriate conversations that move beyond traditional scripted dialogue trees. Recent tool-calling LLMs can both converse naturally and interact with their environment through structured actions, bridging conversational AI with interactive game mechanics \cite{park2023generative}.

The Commonsense Persona-Grounded Dialogue Challenge 2025 required NPCs to demonstrate both conversational competence and the ability to perform actions using available tools within a 7-second response time limit on L40S GPU with 48 GB VRAM. The competition featured three distinct tasks: Task 1 (Task-Oriented Dialogue) focused on agents capable of executing functions and incorporating results into responses; Task 2 (Context-Aware Dialogue) emphasized natural character-appropriate conversation without function execution; and Task 3 (Hybrid) combined both capabilities into a unified system. This dual requirement necessitated a novel architectural approach that could efficiently balance natural language generation with structured tool usage \cite{cpdc2025}.

\section{Model Selection and Rationale}

\subsection{Base Model Selection}
After evaluating multiple candidates, Qwen3 was selected as the base model due to its superior native tool-calling capabilities and well-defined interaction format. Although Qwen3 possesses advanced reasoning capabilities, these were deliberately disabled due to the 7-second time constraint for each NPC turn and Qwen3's tendency to be a lengthy reasoner, which would have exceeded the competition's strict timing requirements. The model outputs tool calls using a structured format:
\begin{verbatim}
<tool_call>
{...well defined json schema...}
</tool_call>
\end{verbatim}
Tool responses are integrated back into the conversation history with clearly defined delimiters:
\begin{verbatim}
<tool_response>
{...well defined json response...}
</tool_response>
\end{verbatim}
This clear formatting simplified training dataset creation and ensured reliable tool interaction patterns \cite{yang2025qwen3, qwen2025blog}.

Training and deployment were facilitated by \href{https://github.com/unslothai/unsloth}{Unsloth}, an optimization library that significantly accelerates fine-tuning and inference of large language models through memory-efficient implementations and kernel optimizations. Unsloth's support for specific model architectures was therefore a critical consideration in model selection, as it directly impacted training feasibility within resource constraints.

\subsection{Alternative Models Considered}
Several other models were evaluated, focusing on those with open, highly permissive licenses from the Berkeley Function-Calling Leaderboard (BFCL) \cite{bfcl_leaderboard_v3_2025}. While other high-performing models existed on the leaderboard, licensing constraints limited viable candidates. \textbf{Table~\ref{tab:model_comparison}} summarizes the compatibility assessment of the considered models.

\begin{table}[H]
\centering
\small
\begin{tabular}{lccc}
\toprule
\textbf{Model} & \textbf{Tool calling} & \textbf{Unsloth} & \textbf{BFCL} \\
 & \textbf{support} & \textbf{support} & \textbf{model size \& rank} \\
\midrule
Qwen3 & FC & \checkmark & 8B \#18 \\
ToolACE2 & FC & \texttimes & 8B \#10 \\
watt-tool & FC & \texttimes & 8B \#15 \\
Gemma3 & Prompt & \checkmark & 12B \#78 \\
Llama3 & FC & \checkmark & 8B \#83 \\
\bottomrule
\end{tabular}
\caption{Model compatibility and BFCL rankings (As of August 14, 2025)}
\label{tab:model_comparison}
\end{table}

FC = native support for function/tool calling. Prompt = workaround for function calling, using model's normal text generation capability.

\begin{itemize}
    \item \textbf{ToolACE-2} \cite{liu2025toolace}: Not supported by Unsloth, limiting training efficiency.
    \item \textbf{watt-tool-8B} (see \href{https://huggingface.co/watt-ai/watt-tool-8B}{link}): Similarly incompatible with Unsloth optimization framework.
\end{itemize}

\begin{itemize}
    \item \textbf{Gemma3} \cite{gemma_2025}: While technically capable of tool calling, it lacked a clearly defined format and was not specifically designed for structured tool usage \cite{google2024gemma-function-calling}, leading to unreliable and inconsistent tool-calling. Consistent with this, the Berkeley Function-Calling Leaderboard shows Gemma3 variants underperforming Qwen3 by a clear margin.
    \item \textbf{Llama3} \cite{llama3_2024}: Tool calling is only available in larger variants (8B/70B/405B). However, the BFCL places Llama3 far behind Qwen3 on tool-calling accuracy.
\end{itemize}

\section{Dataset and Training Methodology}

\subsection{Dataset Preparation}
The training process began with the official challenge datasets (Task 1 and Task 2), provided as JSON files containing conversation lists with multiple turns. Task 3 did not have a dedicated dataset as it was designed as a combination of Tasks 1 and 2. Each turn contained prior conversation context and gold replies with gold function calls. However, the JSON files were not self-contained, requiring additional data retrieval for the specific tools and actions available to the agent at each turn from separate Python files (available at \href{https://github.com/MahammadNuriyev62/CPDC-challenge-2025-solution/blob/main/function_call_langchain}{function\_call\_langchain}). See Appendix~\ref{appendix:dataset_examples} for detailed examples of the original dataset structure and preprocessing pipeline.

The original datasets were distributed through the \code{sony-cpdc-2025-starter-kit} GitLab repository during the competition period. For reference purposes, the training datasets are available at \href{https://github.com/MahammadNuriyev62/CPDC-challenge-2025-solution/blob/main/data/task1_train.json}{task1\_train.json} and \href{https://github.com/MahammadNuriyev62/CPDC-challenge-2025-solution/blob/main/data/task2_train.json}{task2\_train.json}, though these are unofficial copies as the original repository is no longer accessible.

The dataset modification process involved:
\begin{enumerate}
    \item \textbf{Sequential Message Restructuring}: For each conversation, turns were merged into sequential message items (e.g., user message, assistant message, user message, assistant tool call, tool response, assistant message, etc.), with \code{assistant tool call} and \code{tool response} formatted according to Qwen3 recommendations.
    \item \textbf{Tool and Action Schema Integration}: The available tools and actions for each specific turn were retrieved from the Python processing files and converted to appropriate formats (see Appendix~\ref{appendix:conversion_examples} for conversion examples). 
    \item \textbf{Data Augmentation}: A comprehensive three-wave augmentation strategy was implemented to address the limited size of the original training data (detailed methodology provided in Section~\ref{sec:data_augmentation}). GPT-o4-mini-high was used as a data generation tool to create augmented training examples, increasing the dataset size by nearly 300\% with diverse conversational variations while preserving the original tool-calling patterns and persona consistency.
    \item \textbf{Training Split Generation}: During training, conversations were split into input-label pairs at various dialogue points:
    \begin{itemize}
        \item \textbf{input}: user message; \textbf{label}: assistant message
        \item \textbf{input}: user message, assistant message, user message; \textbf{label}: assistant message
        \item \textbf{input}: user message, assistant message, user message; \textbf{label}: assistant tool call
        \item \textbf{input}: user message, assistant tool call, tool response; \textbf{label}: assistant message
        \item etc.
    \end{itemize}
    Note that only the most recent tool calls and tool responses (those immediately preceding the assistant's final reply) were retained in conversation history during training, while earlier tool interactions were removed to prevent the model from overly relying on previous tool outputs that would not be available during actual inference at the current turn, ensuring more robust and generalizable behavior.
\end{enumerate}

\subsection{Data Augmentation Strategy}
\label{sec:data_augmentation}

Given the limited size of the original training datasets, a data augmentation strategy was implemented to improve model performance. GPT-o4-mini-high was employed to generate augmented training examples, leveraging its strong reasoning capabilities to ensure format consistency, accurate tool calling patterns, and conversational coherence in the synthetic data used to train the Qwen3 model variants.

\subsubsection*{Three-Tier Augmentation Approach}

The augmentation process employed three distinct levels of aggressiveness, each targeting different aspects of data variation while preserving the underlying conversation structure:

\begin{itemize} 
\item \textbf{Low Aggressiveness} This conservative approach maintained all structural elements unchanged, focusing solely on linguistic variation. Messages within conversations were paraphrased while preserving exact meaning, tool names, argument structures, and contextual elements.

\item \textbf{Medium Aggressiveness} This intermediate approach preserved the core role and worldview of NPC while introducing moderate structural changes. Tool names, descriptions, arguments, argument descriptions, and conversation phrasing were systematically modified (e.g., \code{check\_amount} became \code{check\_quantity}, \code{check\_price} became \code{lookup\_pricing}).

\item \textbf{High Aggressiveness} This approach introduced substantial changes to roles and worldviews while preserving the fundamental nature of interactions. For example, a ``Merchant in Medieval town'' persona might be transformed into an ``Armor supplier on Mars secret military base.''
\end{itemize}

\subsubsection*{Conversation Flow Preservation and Augmentation Boundaries}

A critical constraint across all augmentation levels was preserving exact conversation flow and interaction patterns. The augmentation process maintained strict 1:1 correspondence in conversation structure, ensuring that functional elements appeared at identical positions across original and augmented dialogues. For instance, if the original conversation contained a price-checking tool call at message \#6, the augmented version would feature an equivalent pricing tool call at the same location. This methodology preserved essential training signals for decision-making patterns while maximizing surface-level diversity. Crucially, role-specific behaviors remained consistent---e.g. selling-related NPCs maintained their commerce focus, and tool usage patterns were systematically preserved across all transformations. "We kept the skeleton the same but changed the skin."

An even more aggressive augmentation approach would have involved generating entirely independent conversation flows, potentially creating more robust models through complete structural diversity. However, this approach was deliberately avoided for two primary reasons: first, to prevent deviation from the provided dataset's inherent patterns and interaction logic, ensuring that augmented data remained faithful to the original challenge requirements; and second, due to concerns that excessive deviation from the original dataset distribution might negatively impact performance on the private evaluation metrics, potentially compromising competitive performance. This methodological decision balanced scientific rigor with practical competition strategy, maintaining the specific conversational dynamics intended by the dataset creators.

\subsubsection*{Prompting Strategy}

Each augmentation request to GPT-o4-mini followed a structured format: ``You have this example, augment it while ensuring the following rules...'' The prompts included specific guidelines for maintaining conversation flow, preserving tool calling patterns, and adhering to the target augmentation level. Reference examples were provided to ensure consistent augmentation quality and format compliance.

It is important to note that GPT-o4-mini served solely as a data generation tool; the final NPC system uses only Qwen3-based models trained on this augmented dataset.

\subsubsection*{Performance Impact Analysis}

The augmentation strategy demonstrated varying effectiveness across model sizes, with smaller models showing more substantial improvements:

\begin{itemize}
    \item \textbf{Qwen3-1.7B}: Performance increased from approximately 0.5 to 0.6+ score (increase by 0.1+ points) on Task 1, representing a significant relative improvement of 20\% or more.
    \item \textbf{Qwen3-14B}: While improvements were less dramatic, performance still increased by approximately 0.030 to 0.040 points on Task 1, demonstrating consistent benefits across model scales.
\end{itemize}

This size-dependent improvement pattern aligns with the hypothesis that smaller models benefit more from increased data diversity due to their limited capacity to generalize from sparse examples, while larger models already possess stronger generalization capabilities.

\subsubsection*{Quality Assurance Framework}

To ensure the integrity of augmented data, a validation framework was developed consisting of three specialized scripts (see the \href{https://github.com/MahammadNuriyev62/CPDC-challenge-2025-solution/tree/main/augmentation#dataset-validation-suite}{Dataset Validation Suite} for code and usage).

\begin{enumerate}
    \item \textbf{Structural Validation}: Verified that message sequences followed expected conversational rules (no consecutive messages from the same speaker), validated XML-like control tags (\code{<tool\_call>}, \code{<tool\_response>}) for proper nesting and closure, and identified exact duplicates or Unicode escape sequence issues.
    
    \item \textbf{Schema Compliance}: Ensured that declared functions adhered to consistent schemas with required fields, validated JSON formatting within \code{<tool\_call>} and \code{<tool\_response>} blocks, and confirmed that function calls referenced existing functions from the declared function list.
    
    \item \textbf{Semantic Integrity}: Verified that provided arguments matched function schemas in both presence and type, ensuring that the augmented conversations maintained logical consistency and executable tool interactions.
\end{enumerate}

This multi-layered validation approach served as a quality gate, preventing malformed message structures, inconsistent metadata, broken tool calls, or schema mismatches from entering the training pipeline. The validation framework was essential for maintaining the reliability of the significantly expanded dataset, which grew by nearly 300\% through the augmentation process (available on Hugging Face at \href{https://huggingface.co/datasets/nuriyev/cpdc-agent}{datasets/nuriyev/cpdc-agent}).

\subsection{Training Strategy}
Due to resource constraints, LoRA (Low-Rank Adaptation) was preferred over full fine-tuning. Initial attempts using a single adapter for both tool calling and natural dialogue resulted in degraded conversational quality. This led to the development of a multi-expert architecture. Importantly, LoRA made this multi-expert approach feasible, as loading several lightweight adapters with a single base model is significantly more VRAM-efficient than deploying multiple full models.

Each of the two LoRA adapters was trained separately on Google Colab using Unsloth. The models were trained for 2 epochs, which typically required 1 to 2 hours on a single A100 GPU. Training and validation losses were monitored to control training improvements and prevent overfitting. Additional metrics were deliberately excluded as they required processing logits, which would have added significant time to training and increased GPU VRAM usage. Instead, the full GPU capabilities were focused on training, with evaluation performed post-training. 

Only the most recent tool calls and tool responses (those immediately preceding the assistant's final reply) are retained in conversation history, while earlier tool interactions are removed to prevent the model from overly relying on previous tool outputs that would not be available during actual inference at the current turn, ensuring more robust and generalizable behavior.

The complete training notebooks for both adapters are publicly available: PersonaLoRA training (\href{https://github.com/MahammadNuriyev62/CPDC-challenge-2025-solution/blob/main/training/Qwen3_persona.ipynb}{Qwen3\_persona.ipynb}) and ToolLoRA training (\href{https://github.com/MahammadNuriyev62/CPDC-challenge-2025-solution/blob/main/training/Qwen3_tool.ipynb}{Qwen3\_tool.ipynb}).

\newcommand{\toolcallprefix}{\code{<tool\_call>\textbackslash n{"name": "}} }

\newcommand{\replytoolcall}{\code{<tool\_call>\textbackslash n{"name": "}} }

\newcommand{\replytoolname}{\code{reply} }

\section{Multi-Expert System Architecture}

\subsection{Three-Expert Design}
The final system employs three specialized components, all publicly available on HuggingFace:

\begin{figure*}[t]
    \centering
    \includegraphics[width=1\textwidth]{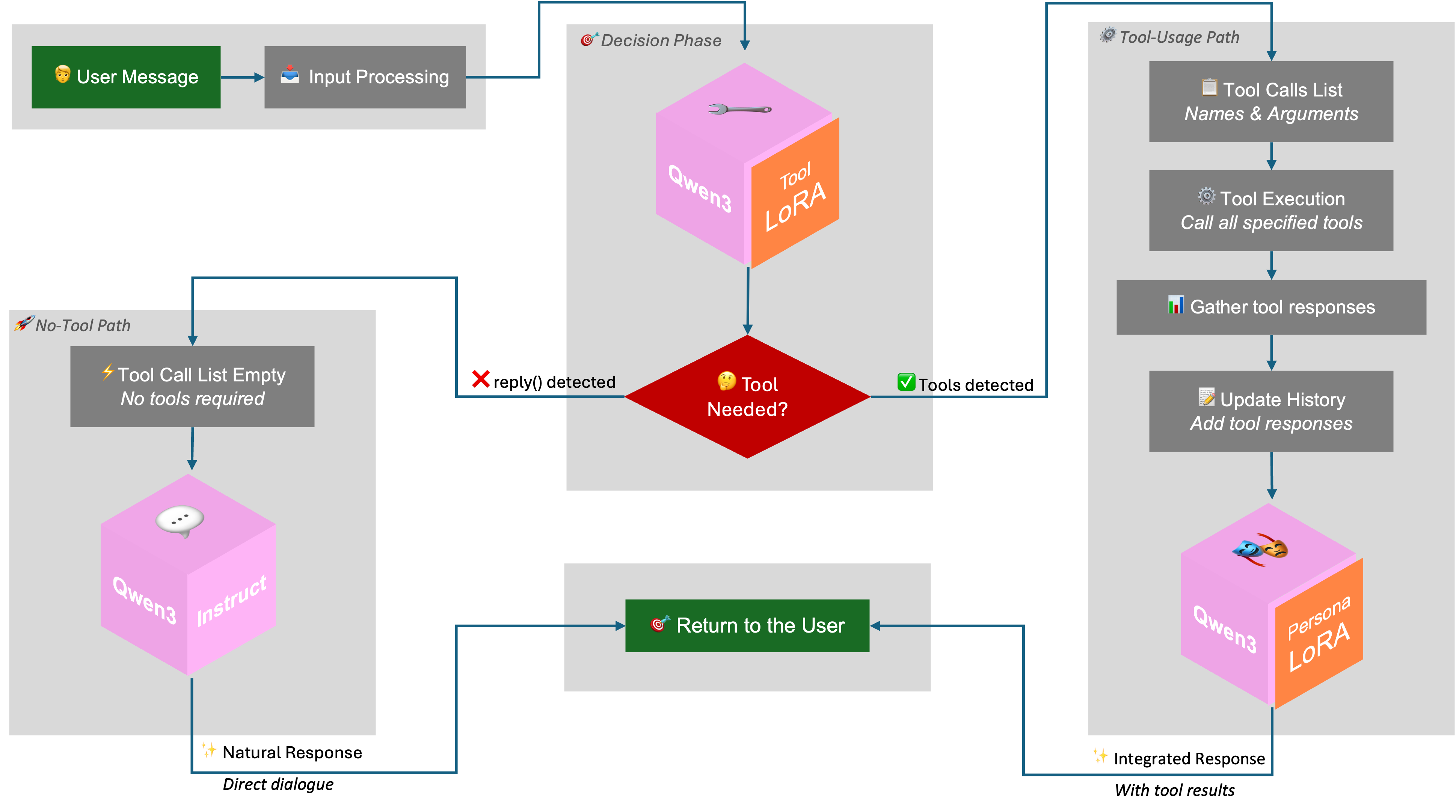}
    \caption{Multi-Expert System Architecture showing the complete inference pipeline with decision phase, tool/no-tool paths, and the three expert models}
    \label{fig:architecture}
\end{figure*}

\paragraph{ToolLoRA (Tool Calling Expert)}

\begin{itemize}
    \item \textbf{Purpose}: Determines whether tool usage is required and generates appropriate tool calls.
    \item \textbf{Special Feature}: Includes a dummy \code{reply()} tool artificially added to the prompt, enabling the model to output \code{\{"name": "reply"\}} when no actual tools are needed.
    \item \textbf{Optimization}: Pre-fills \toolcallprefix during inference (saving 6 tokens per turn) and implements early stopping when \code{reply} is detected (saving an additional 9 tokens). This is possible because ToolLoRA output always starts with \toolcallprefix regardless of the subsequent tool name. Early stopping when \replytoolname is detected eliminates unnecessary token generation since the decision path is already determined. While this approach introduces a potential brittleness risk due to name collision if an actual tool named "reply" existed, this is acceptable in our controlled environment where the tool namespace is managed and such conflicts can be prevented.

\end{itemize}

Available \href{https://huggingface.co/nuriyev/qwen3-14B-cpdc-tool-lora}{here}.

\paragraph{NoLoRA (Direct Reply Expert)}
\begin{itemize}
    \item \textbf{Purpose}: Generates natural responses when no tools are needed, leveraging the original model's conversational capabilities.
    \item \textbf{Base}: Original Qwen3 instruct model without modifications, preserving its natural dialogue generation quality which proved to be very good.
    \item \textbf{Optimization}: Tool lists are deliberately hidden from this model since it is specifically designed to reply naturally when tool calling is not required. Excluding this unnecessary data reduces the input size and makes inference faster, while preventing confusion from irrelevant tool information. Additionally, hard-coded \code{<think>...</think>} tokens with pre-written reasoning content inside are pre-added to force the model's reasoning and guide it toward the desired response pattern.
\end{itemize}

\paragraph{PersonaLoRA (Tool Response Integration Expert)}

\begin{itemize}
    \item \textbf{Purpose}: Interprets tool responses and generates contextually appropriate replies that integrate tool results into natural conversation flow.
    \item \textbf{Training}: Specialized on conversations containing tool interactions, learning to seamlessly incorporate tool outputs into coherent responses.
    \item \textbf{Safeguards}: \code{<tool\_call>} and \code{</tool\_call>} tokens are actively suppressed to eliminate any possibility of accidental tool calling, since this model is specifically designed for response generation rather than tool usage.
    \item \textbf{Guidance}: Hard-coded \code{<think>...</think>} tokens with pre-written reasoning content inside are pre-added to force the model's reasoning and guide it toward the desired response pattern.
\end{itemize}

Available \href{https://huggingface.co/nuriyev/qwen3-14B-cpdc-tool-lora}{here}.

\subsection{Inference Pipeline}

\begin{enumerate}
    \item \textbf{Input Processing}: User message is received by the system.
    \item \textbf{Tool Decision}: User message is routed to ToolLoRA (Qwen3 + tool-calling LoRA adapter) to determine if tools are needed.
    \item \textbf{Branching Decision}:
    \begin{itemize}
        \item If \code{reply() detected}: Early stopping is triggered since no tool call is required, and the user message is routed to NoLoRA (original Qwen3 instruct model) for direct natural response generation.
        \item \textbf{If tool calls detected}: ToolLoRA outputs a list of tool calls with tool names and arguments, proceeding to tool execution phase.
    \end{itemize}
    \item \textbf{Tool Execution}: When actual tool calls are identified, every specified tool is called and their responses are gathered.
    \item \textbf{Response Generation}:
    \begin{itemize}
        \item \textbf{No tools used}: NoLoRA generates the final natural response.
        \item \textbf{Tools used}: Tool responses (including tool names, arguments, and return values) are added to the conversation history and fed to PersonaLoRA (Qwen3 + tool-response interpreting LoRA adapter), which generates the final response incorporating tool results.
    \end{itemize}
    \item \textbf{Output}: Final response from either NoLoRA or PersonaLoRA is returned to the user, completing the current turn.
\end{enumerate}

\section{Performance Results}

\subsection{Competition Performance}
Our system achieved strong performance across all tasks in the CPDC 2025 competition, placing 1st in the public leaderboard and \textbf{2nd overall after human evaluation and final automatic evaluation} (\href{https://discourse.aicrowd.com/t/winners-call-for-paper/17412}{final results}). Table~\ref{tab:competition_results} shows detailed scores across both evaluation phases.

\begin{table}[h]
\centering
\small
\begin{tabular}{lccccc}
\toprule
\multirow{2}{*}{\textbf{Task}} & \multicolumn{2}{c}{\textbf{Public Leaderboard}} & & \multicolumn{2}{c}{\textbf{Final Evaluation}} \\
\cmidrule(lr){2-3} \cmidrule(lr){5-6}
 & \textbf{Score} & \textbf{Rank} & & \textbf{Score} & \textbf{Rank} \\
\midrule
Task 1 & 0.664 & 2nd & & 0.632 & 3rd \\
Task 2 & 0.613 & 1st & & --- & --- \\
\textbf{Task 3} & 0.639 & 1st & & 0.626 & \textbf{2nd} \\
\bottomrule
\end{tabular}
\caption{Competition results for Task 1 (Tool Usage), Task 2 (Dialogue Quality) and Task 3 (Hybrid) across evaluation phases. Final ranking: \textbf{2nd overall}.}
\label{tab:competition_results}
\end{table}

\subsection{Computational Efficiency}
The system exceeds the challenge's computational requirements:
\begin{itemize}
    \item \textbf{Response Time}: Average 3 seconds (57\% under the 7-second limit).
    \item \textbf{Memory Usage}: <30GB VRAM (62.5\% of available 48GB on L40S).
    \item \textbf{Model Scalability}: Testing with Qwen3-1.7B achieved competitive results (0.6 on Task 1), demonstrating flexible size-performance trade-offs. The architecture supports any Qwen3 variant from 0.6B all the way to its largest variant, since all models in the family support the same tool-calling format. This allows users to balance between model size/inference speed and performance based on their specific computational constraints.
\end{itemize}
To maximize performance within computational constraints, Qwen3-14B was selected as the optimal balance between model capability and resource utilization.

\section{Evaluation and Analysis}

\subsection{Evaluation Metrics}
An evaluation framework was developed using a dedicated test dataset created with the same data augmentation technique employed for training. The evaluation system measured multiple dimensions:
\begin{itemize}
    \item \textbf{Tool Call Accuracy}: Analysis of correct and incorrect tool calls, including true positives, false positives, true negatives, and false negatives to assess decision-making quality.
    \item \textbf{Response Similarity}: Comparison between actual generated responses and gold standard responses to measure content quality and appropriateness.
    \item \textbf{Information Integration}: Evaluation of how well tool responses were incorporated into final replies.
\end{itemize}
This evaluation approach enabled identification of specific problems and guided iterative improvements to the current system architecture. The tool call evaluation pipeline is available at \href{https://github.com/MahammadNuriyev62/CPDC-challenge-2025-solution/blob/main/local_run_task1_test.py}{local\_run\_task1\_test.py}.

\subsection{Identified Issues}
Post-training evaluation revealed several challenges, some of which are:
\begin{itemize}
    \item \textbf{Information Omission}: Base Qwen3 occasionally ignored tool responses in generated replies, failing to incorporate important returned information. For example, when asked "What can you tell me about this sword?", the tool call might return comprehensive data (300 Gold, 15 Attack, good for night battles), but the model would mention only the attack value, omitting critical information like price and special features. This issue led to the introduction of PersonaLoRA, which was specifically trained to integrate tool responses into coherent replies.
    \item \textbf{Response Length}: PersonaLoRA demonstrated a tendency toward overly brief responses, potentially reducing the richness of NPC interactions. This was addressed by encouraging longer responses through the forced hard-coded thinking content trick mentioned earlier.
    \item \textbf{Spurious Arguments}: ToolLoRA sometimes generated unnecessary arguments for the \code{reply} tool, producing outputs like \code{arguments: \{"text": "..."\}} instead of the expected empty arguments format. An issue in the dataset regarding misuse of the reply tool was identified and fixed to resolve this problem.
\end{itemize}

\section{Future Work}

\subsection{Knowledge Graph Integration}
A promising avenue for improvement involves building dynamic knowledge graphs using smaller Qwen3 models with batched parallel inference before incorporating the structured knowledge into the NPC prompt \cite{gao2023peacok}. This approach offers several advantages:
\begin{itemize}
    \item \textbf{Input Size Reduction}: By pre-processing and structuring knowledge into graph format, the overall input size to the main inference models can be significantly reduced, leading to faster inference times.
    \item \textbf{Performance Enhancement}: Smaller, more concise knowledge representation should improve model performance by providing better organized and more relevant context.
    \item \textbf{Batched Parallel Processing}: Utilizing smaller Qwen3 models (such as 1.7B or 4B variants) for knowledge graph construction enables efficient batched parallel processing without competing for resources with the main inference pipeline, allowing multiple knowledge queries to be processed simultaneously. This approach is particularly feasible given that our current system uses less than 30GB of the available 48GB VRAM, leaving approximately 18GB of GPU resources available for parallel knowledge processing.
\end{itemize}

\subsection{Constrained Generation}
Current tool calling reliability, while high, could be further improved through structured output constraints:
\begin{itemize}
    \item \textbf{Grammar Enforcement}: Integrate libraries like \href{https://dottxt-ai.github.io/outlines/latest/}{Outlines} \cite{willard2023efficient} or \href{https://github.com/guidance-ai/guidance}{Guidance} \cite{geng2025jsonschemabench} to ensure 100\% reliable tool calls through enforced grammar constraints, completely eliminating malformed tool usage.
    \item \textbf{Structured Output Guarantees}: Enforce JSON schema compliance at the token level, providing mathematical guarantees about output format correctness.
    \item \textbf{Robustness Enhancement}: These constraints would eliminate the current need for invalid call filtering and post-processing, streamlining the inference pipeline while improving reliability.
\end{itemize}

\section{Conclusion}
This work demonstrates that efficient multi-expert architectures can effectively address the dual challenges of natural dialogue and contextual action execution within strict computational constraints. By decomposing the problem into specialized expert modules and leveraging targeted optimizations, we achieved competitive performance while maintaining significant computational headroom.

\section*{Acknowledgments}
We thank the organizers of the Commonsense Persona-Grounded Dialogue Challenge 2025 and the Wordplay Workshop, AIcrowd and Sony for providing this research opportunity. Special appreciation to the Unsloth team for their optimization framework that made efficient training and inference possible, and to the Qwen3 team for developing the foundational model that powered the entire system.

\bibliography{references.bib}

\onecolumn
\appendix
\section{Dataset Structure and Preprocessing Examples}
\label{appendix:dataset_examples}

\subsection{Original JSON Dataset Format}
The original \href{https://github.com/MahammadNuriyev62/CPDC-challenge-2025-solution/blob/main/data/task1_train.json}{task1\_train.json} and \href{https://github.com/MahammadNuriyev62/CPDC-challenge-2025-solution/blob/main/data/task2_train.json}{task2\_train.json} files contained conversations in the following structure:

\footnotesize
\begin{verbatim}
[
  {
    "data_id": "task1_train_0001",
    "total_turn": 7,
    "worldview": "In a world overrun by monsters...",
    "player": {
      "persona": {
        "name": "Lyrien", "age": "32", "gender": "Female", 
        "occupation": "...",
        "appearance": "I am a petite woman with long, curly brown hair...",
        ...
      }
    },
    "npc": {
      "role": "Play the role of a merchant selling weapons...",
      "persona": { 
        "name": "Luna", "age": "57", "gender": "Female",
        "occupation": "Merchant who sells ...",
        ...
      }
    },
    "function_list_id": "function_list_id_0001",
    "state": {
      "datetime": "Summer, 5 PM", "weather": "Rainy", "place": "Weapon shop"
    },
    "knowledge": {
      "knowledge_info": [
        {"name": "Avis Wind", "type": "Bow", "description": "A light ..."},
        ...
      ],
      "general_info": "### Guild and Environment\n- The Guild always..."
    },
    "turn_0": {
      "dialogue": [
        {
          "speaker": "player", "text": "Good evening. I hope I'm not...", 
          "target_item": []
        }
      ],
      "gold_response": "Not at all...",
      "gold_functions": [
        {
          "name": "search_item",
          "parameters": {"item_description": "a more reliable weapon..."},
          "return": [{"information": "many"}]
        }
      ]
    },
    ...
  },
  ...
]
\end{verbatim}
\normalsize

\subsection{Python File Examples}
The \href{https://github.com/MahammadNuriyev62/CPDC-challenge-2025-solution/blob/main/function_call_langchain}{function\_call\_langchain} directory contained separate definitions for tools and actions that had to be retrieved and integrated.

\subsubsection*{Tool Example (from Python files):}
\footnotesize
\begin{verbatim}
# Example tool definition
def check_price(item_name: str) -> List[Dict[str, str]]:
    """
    Check the price of a specified weapon (e.g. Avis Wind, Short Sword, etc.).

    Parameters:
    ----------
    item_name : str
        Specified weapon name (e.g. Avis Wind, Short Sword, etc.). Uses the weapon... 

    Returns:
    -------
    List[Dict[str, str]]
        Outputs the price of the specified weapon (e.g. Avis Wind, Short Sword, etc.)

    """
\end{verbatim}
\normalsize

\subsubsection*{Action Example (from Python files):}

\footnotesize
\begin{verbatim}
# Example action definition  
def equip(item_name: str) -> None:
    """
    Equip the specified weapon (e.g. Avis Wind, Short Sword, etc.).
    
    Parameters:
    ----------
    item_name: str
        Specified weapon name (e.g. Avis Wind, Short Sword, etc.). Uses the weapon...

    Returns:
    -------
    None
    """

    pass
\end{verbatim}
\normalsize

\subsection{Converted Format Examples}
\label{appendix:conversion_examples}
After preprocessing, the tools and actions were integrated into the conversation context in Qwen3-compatible format.

\subsubsection*{Tool Conversion Example:}
\footnotesize
\begin{verbatim}
# Converted format for training
<tools>
{
    "name": "check_price",
    "description": "Check the price of a specified weapon (e.g. Avis Wind...",
    "parameters": {
      "type": "object",
      "properties": {
        "item_name": {
          "type": "string",
          "description": "Specified weapon name (e.g. Avis Wind, Short Sword..."
        }
      }
    }
}
...
</tools>

User: How much is for the Double-Handed sword?
\end{verbatim}
\normalsize

\end{document}